\documentclass[lettersize,journal]{IEEEtran}
\usepackage{amsmath,amsfonts}
\usepackage{algorithmic}
\usepackage{algorithm}
\usepackage{array}
\usepackage{xcolor}
\usepackage{caption}
\usepackage{textcomp}
\usepackage{stfloats}
\usepackage{url}
\usepackage{verbatim}
\usepackage{graphicx}
\usepackage{arydshln} 
\usepackage{booktabs}
\usepackage{multirow}
\usepackage{makecell}
\usepackage{cite}
\usepackage{pifont}
\begin{document}

\title{Deep Semantic Graph Matching for Large-scale Outdoor Point Cloud Registration}

\author{Shaocong Liu, 
        Tao Wang, 
        Yan Zhang, 
        Ruqin Zhou, 
        Li Li, 
        Chenguang Dai, 
        Yongsheng Zhang, 
        Longguang Wang,
        Hanyun Wang
\thanks{This work was supported in part by the National Natural Science Foundation of China (No. 42271457, 62301601), a grant from State Key Laboratory of Resources and Environmental Information System, and a grant from the Program of Song Shan Laboratory (Included in the management of Major Science and Technology Program of Henan Province) (No. 221100211000-02). (Corresponding author: Hanyun Wang.)}
\thanks{Shaocong Liu, Tao Wang, Yan Zhang, Ruqin Zhou, Li Li, Chenguang Dai, Yongsheng Zhang, and Hanyun Wang are with the School of Surveying and Mapping, Information Engineering University, Zhengzhou 450001, China (e-mail: why.scholar@gmail.com).
        
        Longguang Wang is with the University of Air Force, Changchun, China.}
\thanks{Color versions of one or more of the figures in this article are available online at http://ieeexplore.ieee.org.}}

\markboth{Journal of \LaTeX\ Class Files,~Vol.~14, No.~8, August~2021}%
{Shell \MakeLowercase{\textit{et al.}}: A Sample Article Using IEEEtran.cls for IEEE Journals}

\IEEEpubid{0000--0000/00\$00.00~\copyright~2021 IEEE}

\maketitle

\begin{abstract}
Current point cloud registration methods are mainly based on local geometric information and usually ignore the semantic information contained in the scenes. In this paper, we treat the point cloud registration problem as a semantic instance matching and registration task, and propose a deep semantic graph matching method (DeepSGM) for large-scale outdoor point cloud registration. Firstly, the semantic categorical labels of 3D points are obtained using a semantic segmentation network. The adjacent points with the same category labels are then clustered together using the Euclidean clustering algorithm to obtain the semantic instances, which are represented by three kinds of attributes including spatial location information, semantic categorical information, and global geometric shape information. Secondly, the semantic adjacency graph is constructed based on the spatial adjacency relations of semantic instances. To fully explore the topological structures between semantic instances in the same scene and across different scenes, the spatial distribution features and the semantic categorical features are learned with graph convolutional networks, and the global geometric shape features are learned with a PointNet-like network. These three kinds of features are further enhanced with the self-attention and cross-attention mechanisms. Thirdly, the semantic instance matching is formulated as an optimal transport problem, and solved through an optimal matching layer. Finally, the geometric transformation matrix between two point clouds is first estimated by the SVD algorithm and then refined by the ICP algorithm. Experimental results conducted on the KITTI Odometry dataset demonstrate that the proposed method improves the registration performance and outperforms various state-of-the-art methods. 
\end{abstract}


\section{Introduction}
\IEEEPARstart{W}{ith} the development of 3D laser scanning technologies, point cloud data is widely used in various applications, such as {VR/AR, autonomous driving, and smart city \cite{3DSurvey2021}. As an important research topic, point cloud registration aims to align two point clouds obtained from different scenarios into the same coordinate system by calculating a rotation matrix and a translation vector \cite{dong2020registration, zhang2020deep}. 

Existing point cloud registration approaches can be divided into traditional methods and learning-based methods. 
Traditional methods commonly focus on developing handcrafted descriptors for point cloud registration. Specifically, Stein and Medioni \cite{Stein1992} proposed a rough point cloud registration method based on the splash feature descriptor. 
Later, numerous different feature descriptors are designed for point cloud registration, such as Spin Image \cite{spinimages1999}, 3DSC \cite{3DSC2004}, FPFH \cite{FPFH2009}, RoPS \cite{RoPS2013}, B-SHOT \cite{B-SHOT2015}, BSC \cite{BSC2017}, and KDD \cite{KDD2021}. However, handcrafted descriptors rely heavily on expert knowledge and have limited flexibility. Motivated by the great success of deep learning in various areas, many efforts have been made to learn descriptors for point cloud registration in a data-driven manner, including DeepVCP \cite{DeepVCP2019}, DCP \cite{DCP2019}, PointNetLK \cite{PointNetLK2019}, PCRNet \cite{sarode2019pcrnet}, PREDATOR \cite{PREDATOR2021}, GeoTransformer \cite{GeometricTransformer2022}, and BUFFER \cite{Ao_2023_CVPR}.


Recently, as the size of point clouds is scaled up to millions of points, existing learning-based point cloud registration methods suffer considerable computational costs. A straightforward solution is to downsample the large-scale point clouds to smaller ones for higher efficiency. To this end, voxel-grid-based sampling \cite{Point-GNN2020}, Farthest Point Sampling (FPS) \cite{qi2017pointnet++}, Inverse Density Importance Sub-Sampling (IDISS) \cite{groh2018flex}, and random sampling are widely employed. Nevertheless, directly downsampling the point clouds leads to inevitable geometric information loss. To remedy this, keypoint detection methods are proposed to detect keypoints in a point cloud to better preserve the geometric information, such as Kpsnet \cite{du2019kpsnet}, PRNet \cite{wang2019prnet}, DeepVCP \cite{DeepVCP2019} and D3Feat \cite{D3Feat2020}.
However, existing detection methods are mainly based on the local geometric information and rarely explore the semantic categorical information for registration.  

Semantic information has demonstrated its effectiveness in many point cloud research tasks, including scene recognition \cite{SGPR2020}, and closed-loop detection \cite{Gosmatch2020}.
However, semantic information is rarely studied in the task of point cloud registration. Li et al. \cite{li2021automatic} just
utilized the semantic labels for point cloud segmentation and not for the downstream point cloud registration task.  


In order to fully explore the semantic categorical information for point cloud registration, we formulate the point cloud registration problem as a semantic instances matching and registration task, and propose a deep semantic graph matching network (DeepSGM) for large-scale outdoor point cloud registration. Firstly, we obtain the semantic categorical labels of the large-scale outdoor point cloud using an existing semantic segmentation network, and merge the neighboring points with the same semantic labels to obtain semantic instances using the Euclidean clustering algorithm. Each semantic instance is represented by three kinds of attributes including spatial location information, semantic categorical label, and global shape information. Secondly, we construct a semantic adjacency graph based on the spatial adjacency relations of semantic instances. To fully explore the topological structures between semantic instances in the same scene and across two different scenes, the spatial distribution features and the semantic categorical features are learned with graph convolutional networks  (GCN), and the global geometric shape features are learned with PointNet-like network. These three kinds of high-dimensional features are further enhanced with the self-attention and cross-attention mechanisms. Thirdly, we convert the semantic instances matching problem into an optimal transport problem, and the correspondences between semantic instances are obtained through an optimal matching layer. Finally, the transformation matrix is first estimated by the singular value decomposition (SVD) algorithm based on the semantic instance correspondences and then refined by the ICP algorithm.

In summary, our contributions are summarized as follows:
\begin{enumerate}
    \item This paper models the large-scale outdoor point cloud registration problem as a semantic instance matching and registration task and proposes a deep semantic graph matching network for point cloud registration.
    \item To the best of our knowledge, this paper presents the first attempt to use semantic instances instead of the keypoints as the basic units for point cloud registration.    
    \item This paper conducts experiments on large-scale outdoor point clouds and the results demonstrate that the proposed method achieves superior performance compared with the state-of-the-art methods.
\end{enumerate}

\section{Related Work}

In this section, we first briefly review traditional and learning-based point cloud registration methods. Then, we discuss deep graph matching approaches that are related to our work.

\subsection{Traditional Point Cloud Registration}

Traditional point cloud registration methods usually adopt a coarse-to-fine registration strategy to align two partially overlapped point clouds. In the coarse registration stage, RANSAC \cite{fischler1981random} and its variants \cite{barath2018graph} are the dominant approaches to estimate the coarse transformation matrix from point correspondences. For this kind of method, the hand-crafted local feature descriptors are first extracted and then matched through a feature matching algorithm \cite{spinimages1999, 3DSC2004, FPFH2009, RoPS2013}. However, due to its iterative sampling strategy, the RANSAC-based methods are still time-consuming and noise-sensitive. 

Recently, spatial compatibility is widely used in point cloud registration because it considers the spatial consistency between every set of points under rigid transformation \cite{quan2020compatibility, yang2021sac, chen2022sc2, SC2-PCR++, zhang20233d}. This kind of method directly finds correspondence with spatial compatibility between point sets and does not need to extract local feature descriptors for each point. Typical spatial compatibility based methods include 4PCS \cite{aiger20084PCS} and its variants \cite{ge2017automatic, xu2018multiscale, mellado2014super, mohamad2014generalized, mohamad2015super, raposo2017using}, which extract all sets of coplanar 4-points from one point cloud that are approximately congruent to a given planar 4-points under rigid transformation. Chen et al. \cite{chen2022sc2} proposed a second-order spatial compatibility (${SC^2}$) to measure the similarity of two correspondences. The points with the highest ${SC^2}$ scores are selected as seed points and a two-stage sampling strategy is utilized to expand each seed to a consensus set based on the ${SC^2}$ measure matrix. Later, they improved the above method by measuring the quality of the generated models with a new Feature and Spatial consistency constrained Truncated Chamfer Distance (FS-TCD) metric \cite{ SC2-PCR++}. Zhang et al. \cite{zhang20233d} emphasized the local consensus information in a graph and matched each graph node with the appropriate maximal clique. 

\begin{figure*}[t]
\centering
\includegraphics[width=\textwidth]{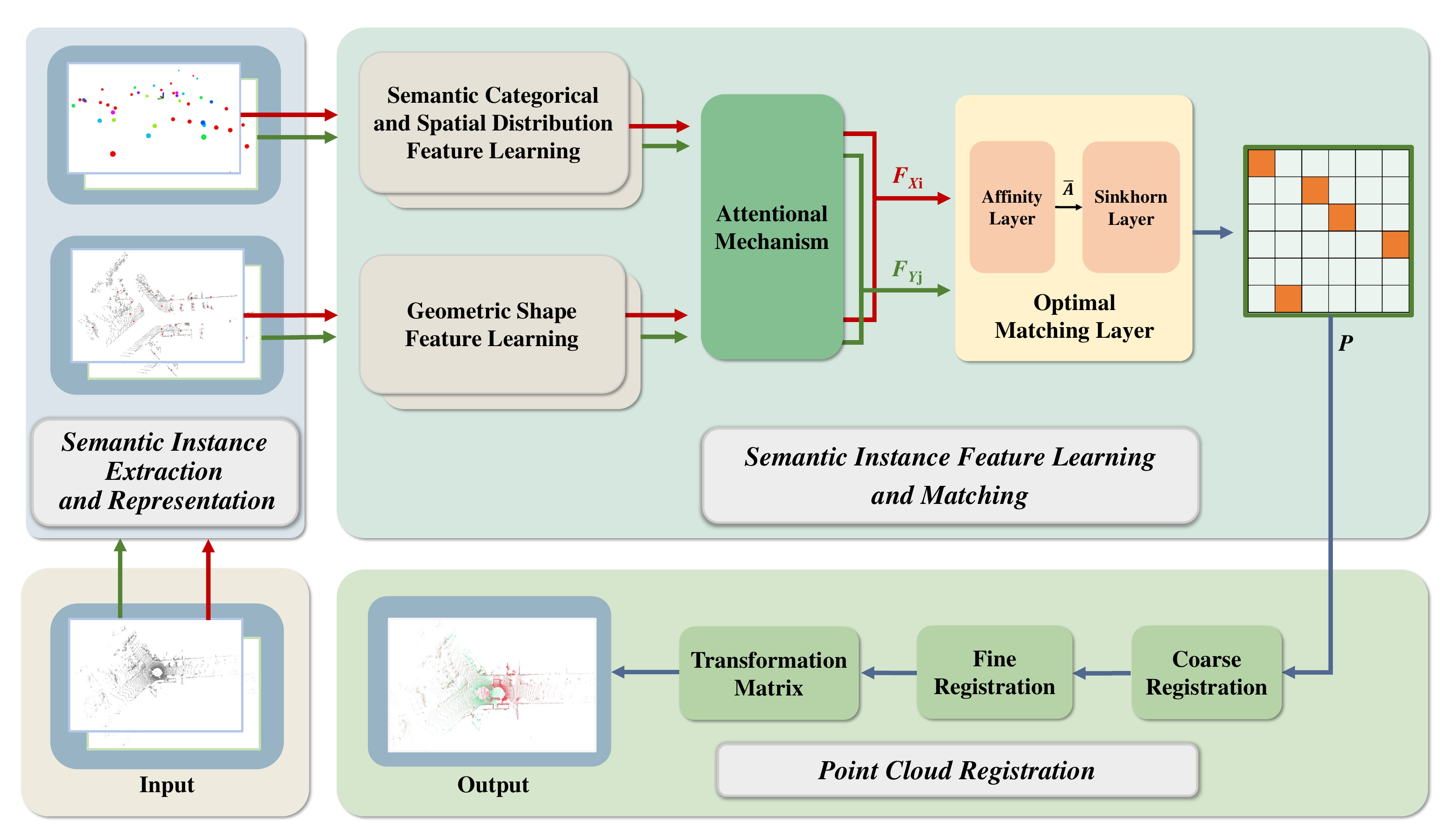}
\caption{The overall pipeline of the proposed method.}
\label{overview}
\end{figure*}

\subsection{Deep Learning-based Point Cloud Registration}

Current deep learning-based point cloud registration methods can be divided into feature matching-based methods, correspondence-based methods, and direct regression methods.

\textbf{Feature matching-based methods.} This kind of method first utilizes the deep neural network to learn pointwise features and then utilizes a traditional feature matching algorithm to obtain point correspondences. Guo et al. \cite{3DSurvey2021} summarized popular 3D point cloud feature learning methods, and divided them into multi-view based methods, volumetric-based, and point-based methods. Choy et al. \cite{choy2019FCGF} adopted the Minkowski Engine \cite{choy20194d} to learn the fully-convolutional geometric features (FCGF) from sparse tensor representation, and evaluated their performance on both the indoor and outdoor point cloud registration tasks. SpinNet \cite{ao2021spinnet} extracts rotationally invariant local features with a spatial point transformer module and a neural feature extractor module. The spatial point transformer module maps the input point cloud into a carefully designed cylindrical space and conducts the point cloud registration with an end-to-end $SO(2)$ optimization. The neural feature extractor module learns the powerful features with both the point-based layers and the 3D cylindrical convolutional layers. PREDATOR \cite{PREDATOR2021} detects the overlap region between two point clouds and samples the interest points based on the output overlapping and matchability scores. Zhou et al. \cite{zhou2022masknet++} proposed an inlier/outlier identification network to predict the overlap region between two point clouds, and then performed point cloud registration based on the identification results.  

\textbf{Keypoints detection-based methods.} PRNet \cite{wang2019prnet} first calculates the saliency scores through the $L_2$ norm of the learned features to extract the keypoints, and then obtains keypoint correspondence through the Gumbel–Softmax Sampler. Kpsnet \cite{du2019kpsnet} takes the points with the highest salient scores as candidate keypoints, and establishes the correspondence through the alignment module. USIP \cite{USIP2019} proposes a feature proposal network to learn the salient uncertainties, and takes the points with the smallest salient uncertainties as the keypoints. D3Feat \cite{D3Feat2020} takes the points with maximal scores across the spatial and channel dimensions as the keypoints, and the keypoint correspondences obtained through nearest neighbors finding in the feature space are used to calculate the transformation matrix. Liu et al. \cite{liu2022rethinking} evaluated four kinds of keypoint detectors based on the D3Feat network \cite{D3Feat2020}, and demonstrated that the MLP-Detector obtains the best keypoint detection and point cloud registration performance on both the indoor and outdoor large-scale datasets. 3D3L \cite{3D3L2021} takes points as the keypoints with larger values in the score map. Yu et al. \cite{yu2021cofinet} first established the point correspondences from the downsampled point clouds and then propagated to the dense point clouds without keypoint detection step. Later, GeoTransformer \cite{GeometricTransformer2022} incorporates the geometric structure into the feature learning stage and obtains point correspondences in a similar manner. Fu et al. \cite{RGM2023} transformed the input point clouds into graphs and established the soft correspondence matrix between points through the deep graph matching method. BUFFER \cite{Ao_2023_CVPR} first utilizes the Point-wise Learner to learn point-wise saliencies and orientations, and then utilizes the Patch-wise Embedder to extract local features and cylindrical feature maps for the selected keypoints. The initial correspondences are then obtained by performing feature matching, and an Inlier Generator is proposed to search inliers to predict the rigid transformation. Given two sets of sparse keypoint correspondences, PointDSC \cite{bai2021pointdsc} removes outlier correspondences by explicitly incorporating spatial consistency into the network and also adopts a hypothesize-and-verify pipeline to obtain an inlier/outlier label for each correspondence.

\textbf{Direct regression-based methods.} PointnetLK \cite{PointNetLK2019} uses PointNet \cite{PointNet2017} and max pooling operation to extract global features of the point cloud, and then iteratively estimates transformation parameters with the improved Lucas Kanade (LK) algorithm. PCRNet \cite{sarode2019pcrnet} feeds the two concatenated global features into the multilayer perceptron (MLP) to estimate the rotation quaternion and three translation parameters directly. 
Lee et al. \cite{lee2021deep} leveraged the Hough voting to directly regress 6D transformation parameter space. OMNet \cite{xu2021omnet} first concatenates the global features extracted from the source and target point clouds by using MLP and max pooling operation, and then obtains the transformation matrix through a rigid transformation regression network. 
SCANet \cite{zhou2021scanet} uses the fully connected layer, spatial attention module, and max pooling layer to obtain the global features of each point, and then uses the channel cross-attention module and the fully connection layer to regress the seven transformation parameters. 

\subsection{Deep Graph Matching}
Graph matching has been discussed in pattern recognition and computer vision for decades \cite{Olivier2011, ufer2017deep, Zhou2016, Minsu2013}. In recent years, research on deep learning-based graph matching has attracted more and more attentions. To preserve the global and local structures of the graph, Nie et al. \cite{Zhi2019} proposed a hyper-clique graph (HCG) matching method by replacing the nodes and hyper-edges with cliques (a set of neighboring nodes in a specific feature space) and hyper-edge linking multiple cliques. Zhang et al. \cite{Zhang2018} proposed two effective optimized algorithms to solve the second-order and high-order graph matching problems. For second-order graph matching, a K-nearest-neighbor-pooling matching method is proposed to integrate feature pooling into graph matching. For high-order graph matching, a sub-pattern structure is introduced for the robust graph matching method. Zanfir and Sminchisescu \cite{Andrei2018} extracted the first-order and second-order features of the graph structure to construct the similarity matrix, and obtained the matching results according to the graph matching algorithm. Wang et al. \cite{Wang2019} proposed an end-to-end deep network pipeline for graph matching by embedding edge information into node feature space.

\begin{figure*}[t]
\centering
\includegraphics[width=.95\textwidth,keepaspectratio]{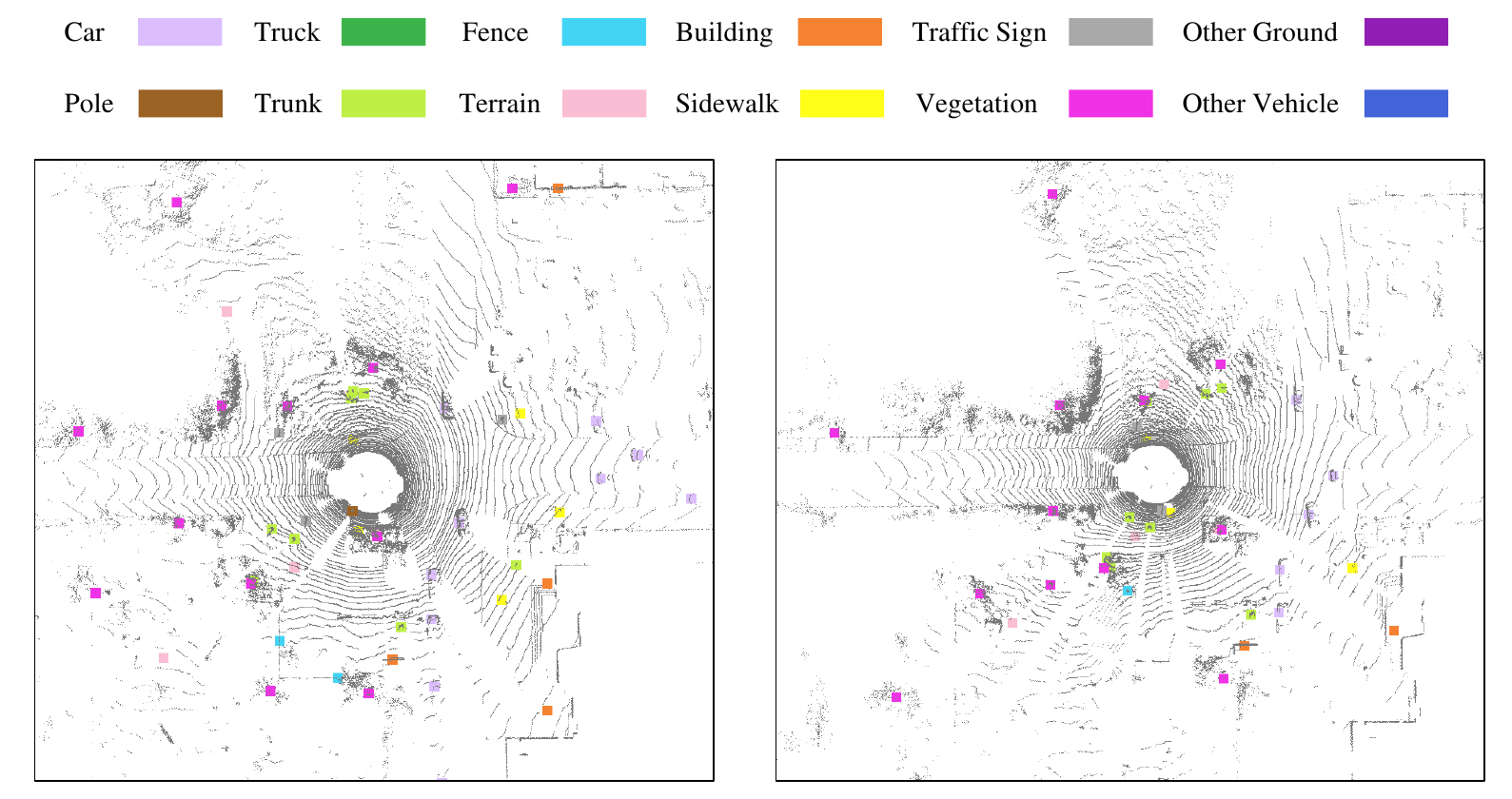}
\caption{The semantic instances extracted from two scenes.}
\label{Fig2_semantic_instance}
\end{figure*}

\section{Methodology}
In this paper, we formulate the point cloud registration problem as a semantic instance matching and registration task, and propose a deep semantic graph matching network (DeepSGM) for large-scale outdoor point cloud registration. The overall pipeline of the proposed method is shown in Fig. \ref{overview}. The proposed network consists of three parts: semantic instance extraction and representation, feature extraction and semantic instance matching, and point cloud registration. The semantic instance extraction and representation module takes the original point clouds as input and uses a semantic segmentation network to extract semantic instances. The feature extraction and semantic instance matching module first constructs the semantic adjacency graph based on the spatial relations between semantic instances and their spatial neighboring instances. To fully explore the topological structures between semantic instances in the same scene and across two different scenes, the spatial distribution features and semantic categorical features are learned through the graph convolutional networks (GCN), and the global geometric shape features are learned through the PointNet-like network. These three kinds of features are then enhanced through the self-attention module performed in the same scene and the cross-attention module performed across two different scenes. In our method, the semantic instance matching problem is formulated as an optimal transport problem, and solved through an optimal matching layer. The point cloud registration module first estimates the coarse rigid transformation matrix with the SVD algorithm by taking the spatial locations of matched semantic instance correspondence as input, and then refines the transformation matrix with the ICP algorithm by taking the original point clouds as input.

\subsection{Semantic Instance Extraction and Representation} \label{sec:extraction and representation}
Inspired by the work \cite{SGPR2020}, semantic instances and their topological relations are critical to identifying the same instances across different scenes. To explore semantic information for point cloud registration, we take semantic instances as the basic processing units instead of the points in the original point clouds. Compared with conventional methods that directly process 3D points, semantic instances are more distinguishable and can also reduce the computational burden. 

Given an original point cloud $P=\{{p_1},{p_2},...,{p_m} |{p_i} \in {R^3}\}$, we first use RangeNet++ \cite{RangeNet++2019} for point cloud semantic segmentation, and obtain semantic instances using Euclidean clustering with different radii for different categorical instances. 
The spatial location of the semantic instance ${I_i}$ is represented by the centroid of the containing points, and is denoted as ${o_i}={({x_{{I_i}}},{y_{{I_i}}},{z_{{I_i}}})^T}$. Since each point in the same semantic instance has the same categorical label, the semantic categorical information of the semantic instance is encoded with a one-hot vector according to the categorical label, and is denoted as ${s_i}={({s_i}^1,{s_i}^2,...,{s_i}^{C})^T}$, where ${C}$ denotes the number of categories in the dataset. The categories with a small number of points (for example, person) are ignored during this process. In addition, in order to obtain the global geometric shape information of each instance, we use the farthest point sampling method to sample $K$ points belonging to each instance from the original point cloud, which is denoted as ${p_i}=\{ {p_{i1}},{p_{i2}},...,{p_{iK}}|{p_{iK}} \in {R^3}\}$. Thus, each semantic instance ${I_i}$ contains three kinds of attributes, including the spatial location information, the semantic categorical information, and the global geometric shape information, which are denoted as ${I_i} = \{ {o_i},{s_i},{p_i}\}$.
Fig. \ref{Fig2_semantic_instance} visualizes the extracted semantic instances of two scenes from the KITTI Odometry dataset, and different colors represent different semantic categories.

\subsection{Feature Learning Module}
\textbf{Semantic instance adjacency graph construction.} 
To fully explore the spatial topological relations of semantic instances, we construct the semantic instance adjacency graph according to the spatial location of semantic instances, as shown in Fig. \ref{Fig3_adjacency_graph}. First, the extracted semantic instances in two large-scale point clouds are taken as the graph nodes and represented as ${I_X} = \{ {I_1},{I_2},...,{I_M}\}$ and ${I_Y} = \{ {I_1},{I_2},...,{I_N}\}$  respectively. Second, according to the spatial location information, the edges ${E_X}$ and ${E_Y}$ of two graphs are constructed by selecting $k$ closet semantic instances for each semantic instance. Two semantic adjacency graphs are finally represented as ${G_X} = \{ {I_X},{E_X}\}$ and ${G_Y} = \{ {I_Y},{E_Y}\}$ respectively.

\begin{figure}[t]
\centering
\includegraphics[width=\columnwidth,keepaspectratio]{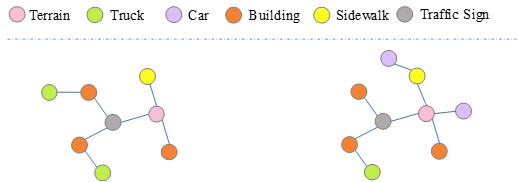}
\caption{Semantic instance adjacency graph based on spatial locations.}
\label{Fig3_adjacency_graph}
\end{figure}

\textbf{Semantic categorical and spatial distribution feature learning.}
Inspired by the DGCNN \cite{wang2019dynamic}, we construct a three-layer GCN with EdgeConv as the core layer to learn the semantic categorical features and spatial distribution features, as shown in Fig. \ref{Fig4_Graph_convolutional_network}.

As described in Section \ref{sec:extraction and representation}, each semantic instance contains spatial location information and semantic categorical information. Since these two attributes describe different characteristics of the semantic instances, we use two different graph convolutional neural networks to learn these two kinds of features respectively. These two networks share the same network structure but with different weights. For the semantic instance ${I_i}$ with spatial location feature ${o_i} = {({x_{{I_i}}},{y_{{I_i}}},{z_{{I_i}}})^T}$ and semantic categorical information ${s_i} = {({s^{1}_i},{s^{2}_i},...,{s^{C}_i})^T}$, $k$ neighborhood semantic instances ${I_{ineigh}} = \{ {I_{i1}},{I_{i2}},...,{I_{ik}} \}$ are first selected according to the edges in the semantic adjacency graph, and the corresponding $k$ spatial location features and semantic categorical features are denoted as ${o_{ineigh}} = {({o_{i1}},{o_{i2}},...,{o_{ik}})^T}$ and ${s_{ineigh}} = {({s_{i1}},{s_{i2}},...,{s_{ik}})^T}$, respectively. Then, the graph convolutional neural networks are used to learn the spatial distribution features and semantic categorical features. Assuming that the input features to the $l$-th layer is ${o^{(l)}_i}$, the output feature of the $l$-th layer can be formulated as follows:

\begin{equation}
{o^{(l + 1)}_i} = \sum\limits_{{o^{(l)}_j} \in {o^{(l)}_{ineigh}}} {{{\varphi}^o _l}} ({o^{(l)}_i},{o^{(l)}_i} - {o^{(l)}_j})
\end{equation}

\begin{equation}
{s^{(l + 1)}_i} = \sum\limits_{{s_j}^{(l)} \in {s^{(l)}_{ineigh}}} {{{\varphi}^s _l}} ({s^{(l)}_i},{s^{(l)}_i} - {s^{(l)}_j})
\end{equation}
where ${{\varphi}^o _l}$ and ${{\varphi}^s _l}$ represent the weighting parameters of the spatial distribution features and semantic
categorical features respectively.

According to the topological relations between the semantic instance and its neighborhood semantic instances, we can obtain two high-dimensional features ${o^{(3)}_i}$ and ${s^{(3)}_i}$ for each semantic instance. 

\begin{figure}[t]
\centering
\includegraphics[width=\columnwidth,keepaspectratio]{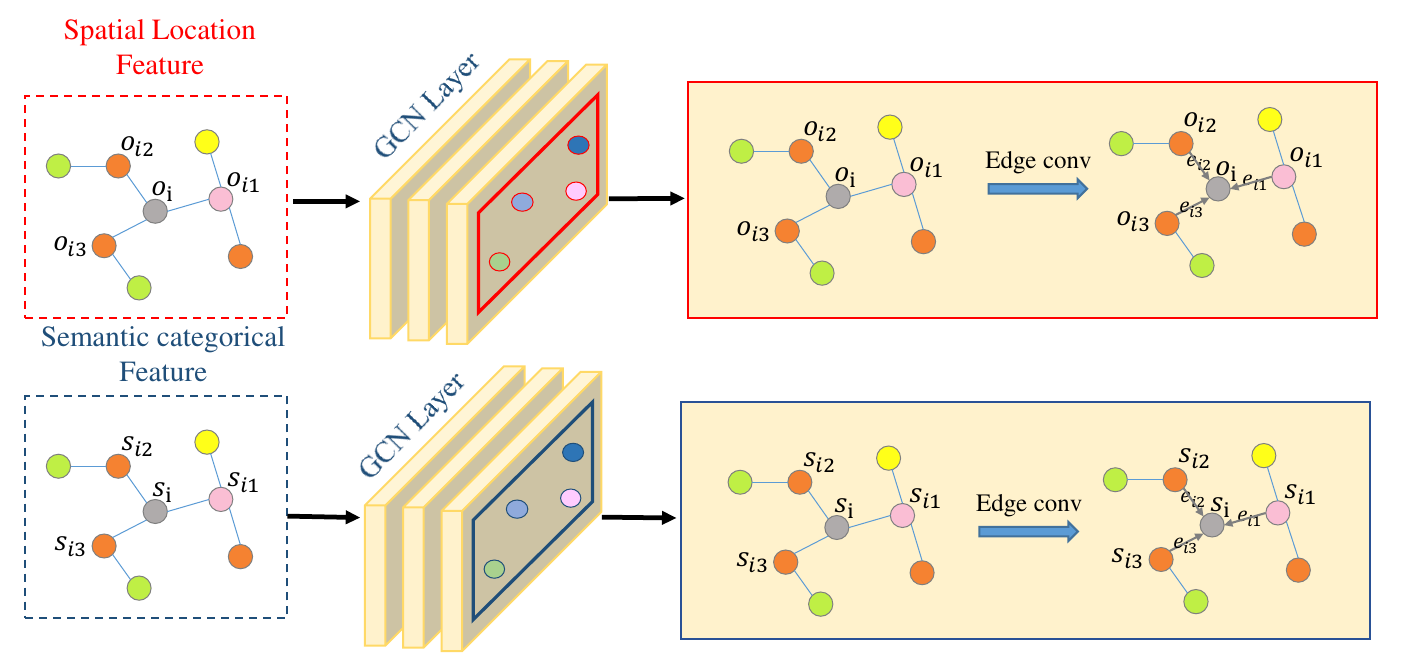}
\caption{Semantic and spatial graph convolutional networks in our method.}
\label{Fig4_Graph_convolutional_network}
\end{figure}

\textbf{Geometric shape feature learning.} 
For semantic instance $I_i$, we sample $K$ points belonging to each instance according to the farthest point sampling method, which is represented as ${p_i} = \{ {p_{i1}},{p_{i2}},...,{p_{iK}}|{p_{iK}} \in R{^3} \}$. Then, we utilize the PointNet-like network to extract global geometric shape features, as shown in Fig. \ref{Fig5_geometrical_shape_feature_extraction}.

The T-Net is utilized to normalize the features of the input point cloud. For $K$ points selected from the semantic instance $I_i$, we utilize the multi-layer perceptron (MLP) to estimate 9 parameters of transformation matrix $H$, and then transform the points belonging to the semantic instance according to $H$, which is formulated as follows:

\begin{equation}
{\widetilde {{p_i}}^T} = H{p_i}^T
\end{equation}

The MLP layer consists of a fully connected layer, a batch normalization function, and a ReLU activation function. The feature learning process is formulated as follows:

\begin{equation}
{h^{(l)}_i} = R({\phi ^{(l)}}({W^{(l)}}{\widetilde {{p_i}}^{(l)}} + {b^{(l)}}))
\end{equation}
where $l=1,2,3$ represents the index of the layers in the MLP, $\phi$ represents the batch normalization function, and $R$ represents the ReLU activation function. Finally, we use the maximum pooling layer to obtain the global geometric shape feature $h_i$ for the semantic instance $I_i$.

\textbf{Attention mechanism.}
To further enhance the feature discrimination ability, we propose a feature enhancement module based on the attention mechanism. Specifically, the self-attention mechanism is utilized to effectively capture the context information from all semantic instances in the same scene, and the cross-attention mechanism is utilized to effectively capture the context information from the semantic instances across two different scenes. The schemas of the self-attention mechanism and the cross-attention mechanism are shown in Fig. \ref{Fig6_attention}.

Here, we take the spatial distribution features as examples to introduce the attention-based feature enhancement module. For two semantic instancess ${I_i}$ and ${I_j}$ in the same scene, we calculate their $q^{s}_{i}$, $k^{s}_{i}$ and $v^{s}_{i}$ values respectively by using the self-attention mechanism according to their spatial distribution features $o^{(3)}_{i}$ and $o^{(3)}_{j}$. The detailed functions are described as follows:

\begin{equation}
{q^{s}_{i}} = W^{s}_{1}o^{(3)}_{i} + b^{s}_{1}
\end{equation}

\begin{equation}
\left[ \begin{array}{l}
{k^{s}_{j}}\\
{v^{s}_{j}}
\end{array} \right] = \left[ \begin{array}{l}
{ W^{s}_{2}}\\
{ W^{s}_{3}}
\end{array} \right]o^{(3)}_{j} + \left[ \begin{array}{l}
{b^{s}_{2}}\\
{b^{s}_{3}}
\end{array} \right]
\end{equation}
where parameters $W^{s}_{i}$ and $b^{s}_{i}$ are shared in the self-attention mechanism for each kind of feature.

\begin{figure}[t]
\centering
\includegraphics[width=\columnwidth,keepaspectratio]{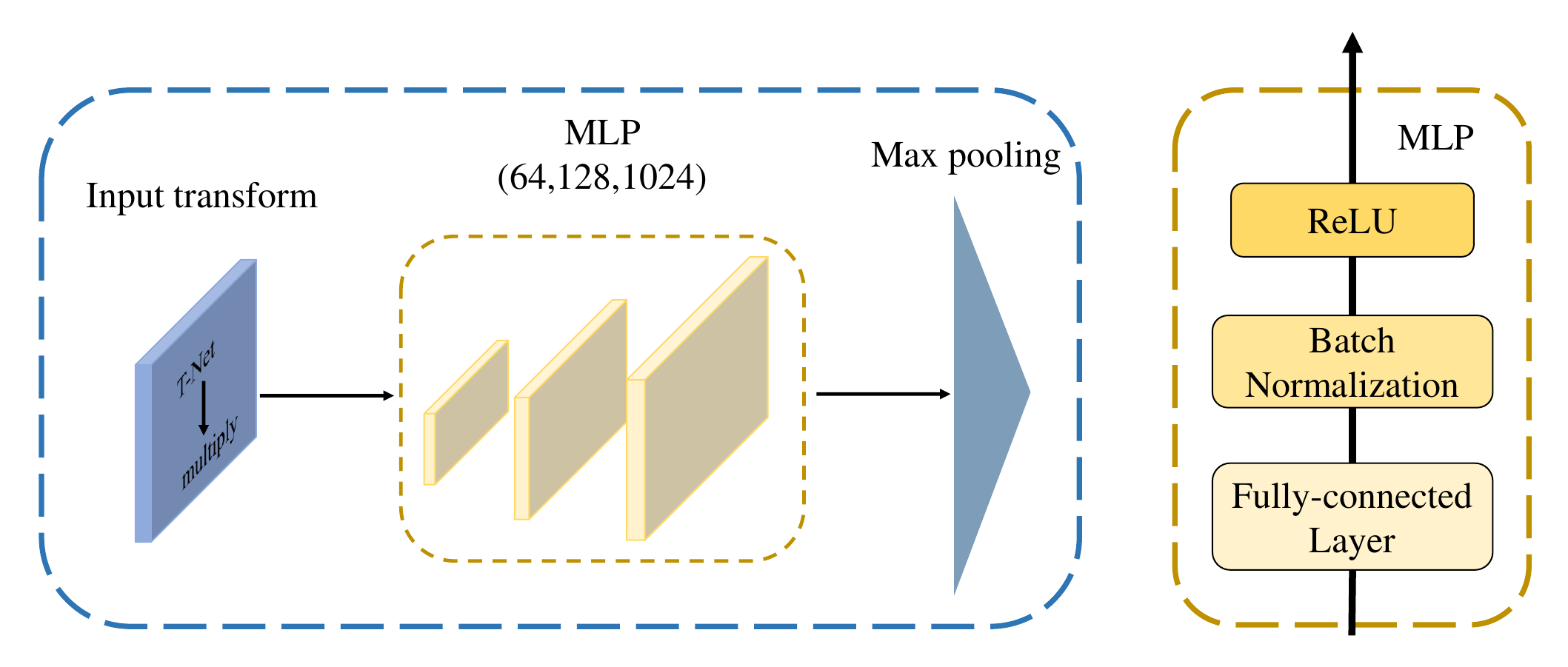}
\caption{the geometrical shape feature extraction network.}
\label{Fig5_geometrical_shape_feature_extraction}
\end{figure}

Then, we calculate the weight score ${\alpha}^{self}_{ij}$ of the spatial distribution features between semantic instance $I_i$ and $I_j$, and the formula is expressed as follows:

\begin{equation}
{\alpha}^{self}_{ij} = {\rm{softmax}}({q^{s}_{i}}^T{k^{s}_{j}})
\end{equation}

Finally, the spatial distribution feature $o^{self}_{i}$ of the semantic instance $I_i$ after the self-attention mechanism can be obtained by the weighted summation of learned features $v^{s}_{i}$ from all other semantic instances in the same scene. The calculation formula is formulated as follows:

\begin{equation}
o^{self}_{i} = \sum\limits_{j = 1}^M {{\alpha}^{self}_{ij}{v^{s}_{j}}}
\end{equation}

For any semantic instance $I_i$ in one scene, the cross-attention mechanism calculates the attention scores of the features between $I_i$ and all semantic instances from another scene, and the final feature is obtained by the weighted summation of features from another scene. Specifically, for semantic instances ${I_i} \in {I_X}$ and ${I_j} \in {I_Y}$ in different scenes, we calculate the $q^{c}_{i}$, $k^{c}_{i}$ and $v^{c}_{i}$ values according to their spatial distribution features $o^{self}_{i}$ and $o^{self}_{j}$, and the formulas are expressed as follows:

\begin{equation}
{q^{c}_{i}} = {W^{c}_{1}}{o^{self}_{i}} + {b^{c}_1}
\end{equation}

\begin{equation}
\left[ \begin{array}{l}
{k^{c}_{j}}\\
{v^{c}_{j}}
\end{array} \right] = \left[ \begin{array}{l}
{W^{c}_{2}}\\
{W^{c}_{3}}
\end{array} \right]{o^{self}_{i}} + \left[ \begin{array}{l}
{b^{c}_{2}}\\
{b^{c}_{3}}
\end{array} \right]
\end{equation}
where parameters $W^{c}_{i}$ and $b^{c}_{i}$ are shared in the cross-attention mechanism for each kind of feature.

Then, we calculate the weight score ${\alpha}^{cross}_{ij}$ of the spatial distribution features between semantic instances ${I_i}$ and ${I_j}$, and obtain the augmented spatial distribution features by the weighted summation of learned features $v^{c}_{i}$ from all semantic instances in another scene according to ${\alpha}^{cross}_{ij}$. The formulas are expressed as follows:

\begin{equation}
{\alpha}^{cross}_{ij} = {\rm{softmax}}({q^{c}_i}^T{k^{c}_{j}})
\end{equation}

\begin{equation}
{o^{cross}_{i}} = \sum\limits_{j = 1}^N {{\alpha}^{cross}_{ij}{v^{c}_j}} 
\end{equation}

In the attention mechanism, the enhancement process of the semantic categorical feature $s$ and the global geometric shape feature $h$ is similar to the spatial distribution feature $o$. After enhancement by the self-attention mechanism and cross-attention mechanism, the high-dimensional features of semantic instance $I_i$ are represented by ${o_i}^{cross}$, ${s_i}^{cross}$ and ${h_i}^{cross}$ respectively. Finally, the feature $F_i$ of the semantic instance $I_i$ is obtained by concatenating the ${o^{cross}_i}$, ${s^{cross}_i}$ and ${h^{cross}_i}$, and is formulated as follows:

\begin{equation}
{F_i} = concat({o^{cross}_i},{s^{cross}_i},{h^{cross}_i})
\end{equation}

\begin{figure}[t]
\centering
\includegraphics[width=0.7\columnwidth,keepaspectratio]{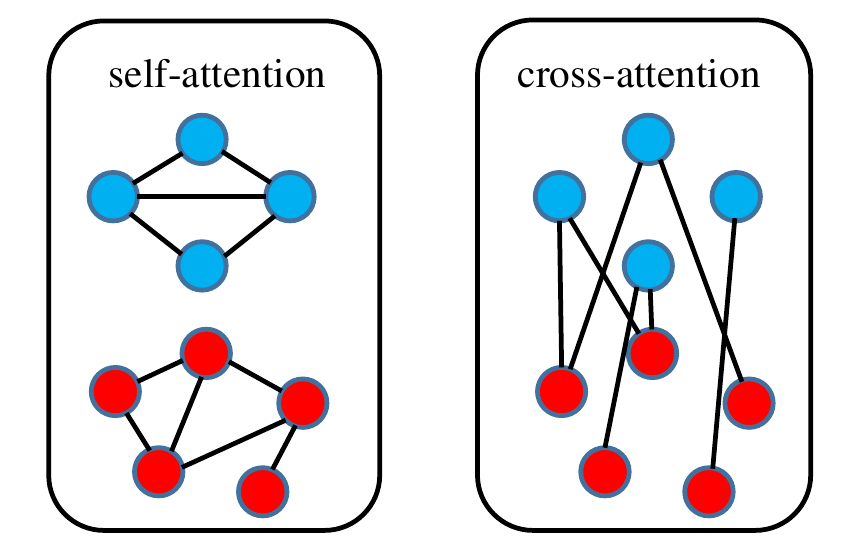}
\caption{The schemas of self-attention mechanism and cross-attention mechanism.}
\label{Fig6_attention}
\end{figure}

\subsection{Optimal Matching Layer}
Recently, the optimal transport model has been widely used in keypoint matching and domain adaption fields, which aims to obtain a mapping function that minimizes the cost of mapping from one distribution to another. Based on the advantage of obtaining the minimum global mapping cost, we use the optimal transport model to implement semantic instance matching. As shown in Fig. \ref{overview}, the proposed optimal matching layer consists of an affinity layer and a Sinkhorn layer.

Given the features ${F_{{I_X}}}$ of the semantic instance set $I_X$ in one point cloud and the features ${F_{{I_Y}}}$ of the semantic instance set $I_Y$ in another point cloud, the affinity matrix ${\rm{A}} \in {R^{M \times N}}$ in the affinity layer is calculated as follows:

\begin{equation}
{\rm{A}} = {({F_{{I_X}}})^{\rm{T}}}W({F_{{I_Y}}})
\end{equation}
where $W$ is the learned parameters. Each element $A_{ij}$ in the affinity matrix $A$ represents the similarity score of two semantic instances ${I_i} \in {I_X}$ and ${I_j} \in {I_Y}$. 

Because the instances in the two point clouds are not always the same, which makes one semantic instance in one point cloud may not have correspondence in another point cloud, similar to Superglue \cite{sarlin2020superglue}, we add a row and a column to the end of the affinity matrix, and is formulated as follows:

\begin{equation}
{\overline {\rm{A}} _{i,N + 1}} = {\overline {\rm{A}} _{M + 1,j}} = {\overline {\rm{A}} _{M + 1,N + 1}} = z \in R
\end{equation}
where $z$ is a learned parameter. The row $M+1$ and column $N+1$ of similarity matrix $\overline {\rm{A}}$ are called dustbins. In this manner, semantic instances without corresponding instances are put into these dustbins.

Then, we use the Sinkhorn algorithm to solve the optimal transport problem and obtain the soft assignment matrix in the Sinkhorn layer. The Sinkhorn algorithm is a differentiable Hungarian algorithm that alternatively normalizes the affinity matrix $\overline {\rm{A}}$ along rows and columns. The details of the Sinkhorn algorithm is shown in Algorithm \ref{alg:alg1}, where $\emptyset$ represents the element-by-element division, and ${{\rm{L}}_M} \in R{^{M \times M}}$ and ${{\rm{L}}_N} \in R{^{N \times N}}$ are identity matrices.

\begin{algorithm}[t]
\caption{Sinkhorn algorithm.}\label{alg:alg1}
\begin{algorithmic}
\STATE \textbf{Inputs:} similarity matrix $\overline {\rm{A}}$, and number of iterations $I$
\STATE \textbf{Outputs:} soft assignment matrix $\overline {\rm{A}}$
\STATE \textbf{Initialization:} ${\overline {\rm{A}} ^{(0)}} = \overline {\rm{A}}$, $k=1$   
\STATE \textbf{while $k<I$ do} 
\STATE \hspace{0.5cm}${\overline {\rm{A}} ^{(k)'}} = {\overline {\rm{A}} ^{(k - 1)}}\emptyset ({\overline {\rm{A}} ^{(k - 1)}}{{\rm{L}}_M})$
\STATE \hspace{0.5cm}${\overline {\rm{A}} ^{(k)}} = {\overline {\rm{A}} ^{(k)'}}\emptyset ({{\rm{L}}_N}{\overline {\rm{A}} ^{(k)'}})$
\STATE \hspace{0.5cm}$k=k+1$
\STATE \textbf{end while}
\STATE $\overline {\rm{P}}  = {\overline {\rm{A}} ^{(k)}}$
\end{algorithmic}
\label{alg1}
\end{algorithm}

Finally, the semantic instance in $I_X$ will be assigned to a single semantic instance in $I_Y$ or the dustbin. After removing the $M+1$ row and $N+1$ column of the matrix $\overline {\rm{P}}$, the remaining matrix $P$ is the soft assignment matrix of the two semantic instance sets from two point clouds. To obtain the hard assignment matrix, two semantic instances are considered as correspondence if the corresponding assignment cost is larger than the predefined threshold $T$.

\subsection{Loss Functions}
According to the soft assignment matrix $P$, we can obtain the semantic instance correspondence set $\Phi \subset {I_X} \times {I_Y}$. To make the obtained semantic instance correspondences close to the ground truth correspondences $\Theta = {\{ (i,j)\}  \subset {I_X} \times {I_Y}}$, the loss function $L$ is defined as follows:

\begin{equation}
L =  - \sum\limits_{(i,j) \in \Theta} {\log {P_{ij}}} 
\end{equation}
and it should meet the following conditions:

\begin{equation}
\forall ({I_{{X_i}}},{I_{{Y_j}}}) \in \Phi, {P_{ij}} > T
\end{equation}
where $P_{ij}$ is the cost of matching semantic instances ${I_{{X_i}}}$ and ${I_{{Y_j}}}$ in the soft assignment matrix $P$, and $T$ is the predefined probability threshold.

\subsection{Registration Based on Semantic Instance Correspondences}
After obtaining the semantic instance correspondences of two inputted point clouds, a coarse-to-fine registration strategy is used to align these two point clouds.

\textbf{Coarse registration.} According to the semantic instance correspondence set $\Phi$ and corresponding spatial locations $O$ of two point clouds obtained in Section \ref{sec:extraction and representation}, we utilize the singular value decomposition (SVD) method to estimate the geometric transformation between two point clouds. Influenced by factors such as occlusions and variations of scanning distances during point cloud acquisition, there exists displacement between the estimated spatial location and the real location of the semantic instance. Therefore, the estimated geometric transformation matrix is not accurate enough and can only be used as coarse registration between two point clouds. 

\textbf{Fine registration.} Based on the coarse geometric transformation matrix ${\{{R_c},{T_c}\}}$ obtained at the coarse registration stage, we adopt the ICP algorithm to further optimize the transformation matrix. Specifically, we take the original point clouds and coarse transformation matrix ${\{{R_c},{T_c}\}}$ as input, and utilize the ICP algorithm to obtain the final transformation matrix ${\{R,T\}}$.


\section{Experiments}
In this section, we first introduce the implementation details, evaluation dataset, and metrics. Then, we compare the performance of our method with previous approaches and conduct ablation experiments to demonstrate our effectiveness.

\subsection{Implementation Details}
To learn the global geometric shape feature for each semantic instance, 128 points are sampled with the FPS algorithm for each semantic instance. At the semantic adjacency graph construction stage, the number of neighboring semantic instances is set to 10, and the output feature dimensions extracted by the graph convolutional neural network are set to 64, 64, and 128, respectively. The batch size, the initial parameters of the momentum optimizer, and the weight decay are set to 32, 0.0001, and 0.98, respectively. The predefined probability threshold $T$ is set to 0.7. All models are implemented using Pytorch and all experiments are conducted on a workstation with an Intel Core i9-9700 CPU, 16GB RAM, and a 24GB NVIDIA Titan GPU. 

\subsection{Evaluation Dataset and Metrics}
\textbf{Evaluation dataset.} 
Different from other point cloud registration methods, the proposed method requires the experimental dataset to provide not only the ground-truth poses but also the pointwise semantic categorical labels. Thus, in this paper we evaluate the performance of the proposed method on the large-scale outdoor point cloud dataset KITTI Odometry. The ground-truth poses are provided by the GPS/INS system, and the pointwise semantic categorical labels are provided in the SemanticKITTI dataset \cite{SemanticKITTI}. In the proposed method, the training set contains 9578 point clouds from 0-5 sequences, the validation set contains 1907 point clouds from 6-7 sequences, and the test set contains 440 point clouds from 8-10 sequences. As shown in Fig. \ref{Fig2_semantic_instance}, only 12 categories are finally reserved in our experiments.

\textbf{Evaluation metrics.} Five evaluation metrics, including the inlier precision (IP), the inlier recall (IR), the relative translation error (RTE), the relative rotation error (RRE), and the registration recall (RR), are adopted for comparison. 

A keypoint correspondence is considered an inlier if its spatial distance transformed according to the ground truth transformation matrix is smaller than a predefined distance threshold. The inlier precision measures the matching quality of the extracted keypoints in the point cloud registration task, while the inlier recall measures the retrieving ability of ground-truth keypoints by the proposed method. In this paper, we take the semantic instance as the basic processing unit, and define the inlier precision  metric $IP$ and the inlier recall metric $IR$ as follows: 

\begin{equation}
IP = \frac{1}{{\left| \Omega  \right|}}\sum\limits_{\left( {{o_{{x_i}}},{o_{{y_j}}}} \right) \in \Omega } {1\left( {\left\| {({R_g}{o_{{x_i}}} + {T_g}) - {o_{{y_j}}}} \right\| < \beta } \right)}
\end{equation}

\begin{equation}
IR = \frac{1}{{\left| {{\Omega}^*}  \right|}}\sum\limits_{\left( {{o^*_{x_i}},{o^*_{y_j}}} \right) \in {{\Omega}^*} } {1\left( {\left\| {({R_g}{o^*_{x_i}} + {T_g}) - {o^*_{y_j}}} \right\| < \beta } \right)}
\end{equation}

where ${{\Omega}^*}$ and $\Omega$ are the spatial location sets of the ground-truth semantic instance correspondences and the predicted semantic instances correspondences respectively. ${o^*_{x_i}}$ and ${o^*_{y_j}}$ are the spatial locations of the ground-truth semantic instance correspondences belonging to the set ${\Omega}^*$, while 
 ${o_{x_i}}$ and ${o_{y_j}}$ are the spatial locations of the predicted semantic instance correspondences belonging to the set ${\Omega}$. The $1{\rm{(}} \cdot {\rm{)}}$ is the counting function, and ${R_g}$ and ${T_g}$ represent the ground-truth rotation matrix and translation vector contained in KITTI Odometry dataset, respectively. The $\beta$ is the distance threshold and is set to 1m in our method.

The RRE and RTE are calculated as follows:

\begin{equation}
{RRE} = \sum\limits_{i = 1}^3 {\left| {angle(i)} \right|}
\end{equation}

\begin{equation}
angle = {\zeta}({R_g}^{ - 1}R)
\end{equation}

\begin{equation}
{RTE} = {\left\| {{T_g} - T} \right\|_2}
\end{equation}
where $R$ and $T$ are the estimated rotation matrix and the translation vector, the function ${\zeta}( \cdot )$ converts the rotation matrix into three Euler angles. 

The registration recall (RR) refers to the fraction of the point cloud pairs whose RRE and RTE are both below predefined thresholds. 
\begin{equation}
RR = \frac{1}{{\Psi}}\sum\limits_{i = 1}^{\Psi} {1( {{RRE_i} < t_{RRE}} \And {{RTE_i} < t_{RTE}})}
\end{equation}
where ${\Psi}$ is the number of point cloud pairs that can be successfully aligned according to the ground-truth transformation matrix. The $t_{RRE}$ and $t_{RTE}$ are the predefined RRE threshold and RTE threshold, which are set to  $5^{\circ}$ and 2m respectively in our experiments.








\subsection{Evaluation Results on the KITTI Odometry Dataset}

\begin{table}[t]
\caption{Registration performance on KITTI Odometry. \label{tab:registration}}
\centering
\renewcommand{\arraystretch}{1.3}
\begin{tabular}{llllll}
\toprule
\multirow{2}{*}{Methods} & \multicolumn{2}{c}{RTE (cm)}         & \multicolumn{2}{c}{RRE (deg)}     & \multirow{2}{*}{RR (\%)}    \\  
                         & \multicolumn{1}{c}{AVG}  & STD  & \multicolumn{1}{c}{AVG}  & STD   \\ \hline
                         
3DFeatNet \cite{yew20183dfeat}   & \multicolumn{1}{c}{25.9} & 26.2 & \multicolumn{1}{c}{0.57} & 0.46 & 96.0\\ 

SKD+3DFeatNet \cite{tinchev2021skd} & \multicolumn{1}{c}{14.0} & 13.4 & \multicolumn{1}{c}{0.57} & 0.48 & 96.52 \\ 

FCGF \cite{choy2019FCGF}  & \multicolumn{1}{c}{9.52} & 1.30  & \multicolumn{1}{c}{0.30}  & 0.28 & 96.6 \\ 

D3Feat \cite{D3Feat2020}  & \multicolumn{1}{c}{6.90}  & 0.30  & \multicolumn{1}{c}{0.24} & \textbf{0.06} & \textbf{99.8} \\ 


PREDATOR \cite{PREDATOR2021}  & \multicolumn{1}{c}{6.80}  & -  & \multicolumn{1}{c}{0.27} & - & 99.8 \\

HRegNet \cite{lu2021hregnet}  & \multicolumn{1}{c}{12.0}  & 13.0  & \multicolumn{1}{c}{0.29} & 0.25 & 99.7 \\

PointDSC+FPFH \cite{bai2021pointdsc}  & \multicolumn{1}{c}{8.13}  & -  & \multicolumn{1}{c}{0.35} & - & 98.20 \\ 

PointDSC+FCGF \cite{bai2021pointdsc}  & \multicolumn{1}{c}{20.94}  & -  & \multicolumn{1}{c}{0.33} & -  & 98.20 \\ 

GeoTransformer \cite{GeometricTransformer2022}     & \multicolumn{1}{c}{6.80}  & -  & \multicolumn{1}{c}{0.24} & -  & 99.8 \\ 

MAC+FPFH \cite{zhang20233d}     & \multicolumn{1}{c}{8.46}  & -  & \multicolumn{1}{c}{0.40} & - & 99.46 \\ 

MAC+FCGF \cite{zhang20233d}     & \multicolumn{1}{c}{19.34}  & -  & \multicolumn{1}{c}{0.34} & - & 97.84 \\ 

DeepSGM                     & \multicolumn{1}{c}{\textbf{6.60}} & \textbf{0.13} & \multicolumn{1}{c}{\textbf{0.229}} & 0.38 & 90.68 \\ 

\bottomrule
\end{tabular}
\end{table}


\begin{table*}[t]
\caption{The detailed configurations of the ablation experiments. \label{tab:alation_models}}
\centering
\renewcommand{\arraystretch}{1.3}
\resizebox{.9\textwidth}{!}{
\begin{tabular}{ccccccc} 


\toprule
Model &  Spatial location  & Semantic category & Geometric shape & PointNet & Self-attention & Cross-attention \\ 
\hline




DeepSGM-1 & \checkmark & \checkmark &  &  & \checkmark & \checkmark  \\

DeepSGM-2 &  & \checkmark & \checkmark & \checkmark & \checkmark & \checkmark  \\

DeepSGM-3 & \checkmark & \checkmark & \checkmark & \checkmark &  & \checkmark  \\

DeepSGM-4 & \checkmark & \checkmark & \checkmark & \checkmark & \checkmark &   \\

DeepSGM & \checkmark & \checkmark & \checkmark & \checkmark & \checkmark & \checkmark  \\
\bottomrule
\end{tabular}
}
\end{table*}

\begin{table*}[t]
\caption{The analytical results of ablation experiments. \label{tab:alation_results}}
\centering
\renewcommand{\arraystretch}{1.3}
\resizebox{.9\textwidth}{!}{
\begin{tabular}{cccccccc} 


\toprule
\multicolumn{1}{c}{\multirow{2}{*}{Methods}} & \multicolumn{2}{c}{RTE(cm)}                        & \multicolumn{2}{c}{RRE(°)}                           & \multicolumn{1}{c}{\multirow{2}{*}{Registration Recall (\%)}} & \multicolumn{1}{c}{\multirow{2}{*}{Inlier Recall (\%)}} & \multicolumn{1}{c}{\multirow{2}{*}{Inlier Precision (\%)}}\\
\multicolumn{1}{c}{}   & \multicolumn{1}{c}{AVG} & \multicolumn{1}{c}{STD} & \multicolumn{1}{c}{AVG}   & \multicolumn{1}{c}{STD} & \multicolumn{1}{c}{}  & \multicolumn{1}{c}{}  & \multicolumn{1}{c}{}  \\ \hline

DeepSGM-1  & 6.8 & 0.13   & 0.366 & 0.74  & 77.04  &62.5  & 82.32      \\ 

DeepSGM-2  & 7.3 & 0.16   & 0.25  & 0.40  & 87.73  &59.63  & 89.28      \\ 

DeepSGM-3  & 6.6 & 0.13   & 0.24  & 0.37  & 85.91  &48.66  & 90.7       \\ 

DeepSGM-4  & 6.7 & 0.14   & 0.248 & 0.39  & 84.32  &43.32  & 90.91      \\ 

DeepSGM    & 6.6 & 0.13   & 0.229 & 0.38  & 90.68  &54.25  & 92.97      \\ 
\bottomrule

\end{tabular}
}
\end{table*}

Following the state-of-the-art methods, we use the RTE, RRE and RR to evaluate the registration performance on the KITTI Odometry Datase. The following methods are compared with the proposed DeepSGM method, including 3DFeatNet \cite{yew20183dfeat}, SKD \cite{tinchev2021skd}, FCGF \cite{choy2019FCGF}, D3Feat \cite{D3Feat2020}, PREDATOR \cite{PREDATOR2021}, HRegNet \cite{lu2021hregnet}, PointDSC \cite{bai2021pointdsc}, GeoTransformer \cite{GeometricTransformer2022}, and MAC \cite{zhang20233d}. 

As shown in Table \ref{tab:registration}, the DeepSGM method achieves the best RRE and RTE compared with all other methods on the KITTI Odometry dataset. The average RTE and the standard deviation of RTE obtained by DeepSGM are 6.13cm and 0.08cm respectively, and the average RRE and the standard deviation of RRE obtained by DeepSGM are 0.22° and  0.27° respectively. The reasons for the better performance of the DeepSGM lie in three aspects. First, the DeepSGM takes the semantic instances as the matched correspondences instead of 3D points. The performance of methods that take 3D points as correspondences is easily affected by similar geometric structures in the point cloud. However, taking the semantic instances as correspondences is more robust compared with 3D points. Second, the feature learning module in the DeepSGM utilizes three kinds of attributes to describe the semantic instances and augments the learned features by exploring the spatial topological relationship between semantic instances from two point clouds. Third, a coarse-to-fine registration strategy is adopted in the DeepSGM. The transformation matrix is first estimated based on the spatial locations of the semantic instance correspondences and then refined with the ICP algorithm. However, the registration recall of the DeepSGM is relatively lower than other methods. This is because the number of semantic instances in some point clouds is too little to be used for registration, and we remove these point clouds in the registration process. 



\subsection{Ablation Study}
In this section, we conduct the ablation experiments to analyze the effectiveness of the attributes of the semantic instances and the modules of the network respectively. 

To evaluate the effectiveness of the attributes of the semantic instances in the feature learning stage, we remove the geometric shape information and the spatial location information respectively, and denote the corresponding models as DeepSGM-1 and DeepSGM-2. Because the global geometric shape features are learned from geometric shape information, we correspondingly remove the PointNet-like module in the DeepSGM-1 model. To evaluate the effectiveness of the modules of the network, we remove the self-attention mechanism and the cross-attention mechanism respectively, and denote the corresponding models as DeepSGM-3 and DeepSGM-4. The detailed configurations of the ablation experiments are shown in Table \ref{tab:alation_models}.

Except for the RTE, RRE and RR, inlier precision and inlier recall are also adopted in this section. 
The results of the ablation experiments are shown in Table \ref{tab:alation_results}. Compared DeepSGM-1 and DeepSGM-2 with DeepSGM, we can observe that learning with the geometric shape information and the spatial location information can improve the large-scale outdoor point cloud registration performance. From the gains of the registration success rate and inlier precision, we can observe that the geometric shape information performs more important effectiveness than the spatial location information in the feature learning stage. However, because of the similar instances in the scene, incorporating the global geometric shape information decreases the inlier recall. 
Compared DeepSGM-3 and DeepSGM-4 with DeepSGM, we can also observe that both the self-attention and the cross-attention mechanisms can boost the registration performance. The main effect of the attention mechanism is on the registration success rate, and it promotes at least 5\% success rate in the KITTI Odometry dataset. 


\section{Conclusion}
In this paper, we study the large-scale outdoor point cloud registration problem and propose a deep semantic graph matching method for semantic instance matching and registration. Three kinds of attributes including global geometric shape information, semantic categorical information, and spatial location information are utilized to describe the semantic instances. The high-dimensional features are learned for each kind of feature through the graph convolutional networks and PointNet-like network, and then enhanced based on self-attention and cross-attention mechanisms. The concatenated features are then fed into the optimal matching layer to obtain semantic instance correspondences. The final geometric transformation matrix is then estimated through a coarse-to-fine registration strategy. The experimental results conducted on the KITTI Odometry dataset demonstrate the feasibility and superiority of the proposed method.

\vfill

\end{document}